\documentclass{article}

\PassOptionsToPackage{numbers,sort&compress}{natbib}
\usepackage[sglblindworkshop]{neurips_2025}
\workshoptitle{AICC - AI for Climate and Conservation}

\usepackage[utf8]{inputenc}
\usepackage[T1]{fontenc}
\usepackage{hyperref}
\usepackage{url}
\usepackage{booktabs}
\usepackage{amsfonts}
\usepackage{nicefrac}
\usepackage{microtype}
\usepackage{xcolor}
\usepackage{amsmath}
\usepackage{amssymb}
\usepackage{graphicx}

\title{Zero-Shot Wildlife Sorting Using Vision Transformers 
Evaluating Clustering and Continuous Similarity Ordering}

\author{%
  Hugo Markoff$^{*}$ \\
  CTO Animal Detect \\
  Animal Detect \\
  Aalborg, Denmark \\
  \texttt{hugo@animaldetect.com} \\
  \And
  Jevgenijs Galaktionovs \\
  CEO Animal Detect \\
  Animal Detect \\
  Aalborg, Denmark \\
  \texttt{eugene@animaldetect.com} \\
}

\usepackage{etoolbox}

\AtBeginDocument{\nolinenumbers}

\begin{document}

\maketitle

\noindent $^{*}$Corresponding author

\begin{abstract}
Camera traps generate millions of wildlife images, yet many datasets contain 
species absent from existing classifiers. 
This work evaluates zero-shot approaches for organizing unlabeled wildlife 
imagery using self-supervised vision transformers, developed and tested within 
the Animal Detect platform~\cite{animaldetect} for camera trap analysis. 
We compare unsupervised clustering methods (DBSCAN, GMM) across three 
architectures (CLIP, DINOv2, MegaDescriptor) combined with dimensionality 
reduction techniques (PCA, UMAP), and demonstrate continuous 1D similarity 
ordering via t-SNE projection. 
On a 5-species test set with ground truth labels used only for evaluation, 
DINOv2 with UMAP and GMM achieves 88.6\% accuracy (macro-F1=0.874), while 1D sorting reaches 88.2\% 
coherence for mammals/birds and 95.2\% for fish across 1,500 images. 
Based on these findings, we deployed continuous similarity ordering in 
production, enabling rapid exploratory analysis and accelerating manual 
annotation workflows for biodiversity monitoring.
\end{abstract}

\section{Introduction}

Camera traps worldwide generate millions of wildlife images annually, 
creating annotation bottlenecks \cite{glover2019camera}. 
While convolutional neural network-based classifiers can achieve high accuracy 
when training data covers target species, scenarios when there is a significant domain 
shift or unknown species remain challenging.
Self-supervised vision transformers such as DINOv2 \cite{oquab2023dinov2} 
and language-supervised CLIP \cite{radford2021learning} offer potential solutions through learned 
visual similarity representations that can be clustered without requiring 
species-labeled training data.

This work evaluates whether zero-shot clustering and continuous similarity 
ordering can organize wildlife imagery for conservation workflows within 
the Animal Detect platform~\cite{animaldetect}.

Existing wildlife processing platforms such as Wildlife Insights~\cite{wildlifeinsights2020}, Trapper~\cite{trapper2016}, 
and others rely on species classifiers trained on predetermined 
taxonomies, limiting their applicability when encountering new species. 
Moreover, these systems collapse biological diversity into discrete species 
labels, discarding fine-grained intra-species variation (sex, age, individual 
identity) that is often critical for ecological research and population monitoring.

We investigate zero-shot similarity ordering approaches that (1) operate 
without requiring species-specific training data and (2) preserve continuous 
morphological structure within the sorted sequence, allowing users to discover 
and annotate sub-species patterns beyond simple classification. 
We compare discrete clustering methods with continuous 1D similarity sorting, 
evaluating their potential to accelerate biodiversity monitoring workflows 
in climate-impacted ecosystems.

\section{Methods}

\subsection{Pipeline and Datasets}

All experiments employ a two-stage pipeline: (1) domain-specific detectors 
(MegaDetector v5a \cite{beery2019efficient}~\citep{megadetector_github} for terrestrial, MegaFishDetector~\citep{megafishdetector} 
for aquatic), (2) vision transformer feature extraction from detected crops.

\textbf{Datasets.} Table~\ref{tab:datasets} summarizes the three evaluation 
datasets used in this study.

\begin{table}[h]
\centering
\caption{Evaluation datasets}
\label{tab:datasets}
\small
\begin{tabular}{@{}llcl@{}}
\toprule
\textbf{Dataset} & \textbf{Location/Domain} & \textbf{Images} & \textbf{Species (n=5 each)} \\
\midrule
Vejlerne & Denmark wetlands & 500 & 
  Badger, raccoon dog, red fox, polecat, hooded crow \\
\midrule
Desert Lion \cite{lila2019desert} & African savanna & 500 & 
  Lion, pied crow, ostrich, oryx, giraffe \\
\midrule
DeepFish \cite{saleh2020deepfish} & Tropical reef & 500 & 
  Longfin batfish, sixbar wrasse, grouper (genus), \\
  & & & barramundi, great barracuda \\
\bottomrule
\end{tabular}
\end{table}

\subsection{Clustering Experiments}

We test combinations of three vision transformers (CLIP ViT~\citep{dosovitskiy2021an}-L/14, 
DINOv2 ViT-G/14, MegaDescriptor-L-384) with unsupervised clustering algorithms 
(DBSCAN, GMM) after dimensionality reduction (PCA, UMAP). 
All methods operate without access to species labels; ground truth annotations 
are used solely for post-hoc evaluation.

For Gaussian Mixture Models, we determine the optimal number of components k 
using the Bayesian Information Criterion (BIC):
\begin{equation}
\text{BIC}(k) = -2 \ln(\mathcal{L}) + p(k) \ln(N)
\end{equation}
where \(\mathcal{L}\) is the likelihood, \(N\) is the number of samples, and 
\(p(k)\) is the total number of free parameters. For GMMs with full covariance 
matrices in \(d\) dimensions, \(p(k) = k[d + \frac{d(d+1)}{2}] + (k-1)\) 
(accounting for means, covariances, and mixture weights).
We select \(k^* = \arg\min_{k \in [2, 15]} \text{BIC}(k)\). 
The upper bound of 15 was chosen conservatively based on typical camera trap 
deployments containing limited species per site; for datasets with more species, 
this range should be expanded accordingly.

For each species \(s\), we compute precision/recall using true positives 
(correctly assigned), false positives (other species in cluster), 
false negatives (species in wrong clusters):
\begin{equation}
\text{F1}_s = 2 \cdot \frac{\text{Precision}_s \cdot \text{Recall}_s}
{\text{Precision}_s + \text{Recall}_s}, 
\quad \text{F1}_{\text{macro}} = \frac{1}{5}\sum_{s=1}^{5} \text{F1}_s
\end{equation}
\subsection{Continuous 1D Similarity Ordering}

Rather than discrete clusters, we project embeddings 
\(\{\mathbf{e}_i\}_{i=1}^N\) to 1D using t-SNE (perplexity=30) and sort 
by position. 
Coherence measures the longest continuous species run:
\begin{equation}
\text{Coherence}_s = \frac{\text{max run length of species } s}
{\text{total count of species } s} \times 100\%
\end{equation}
We report mean ± std over 10 independent runs due to t-SNE stochasticity.

\section{Results}

\subsection{Clustering Performance}

DINOv2 ViT-G/14 with UMAP dimensionality reduction followed by GMM 
clustering achieved best performance: BIC selected exactly 5 
components, with \textbf{443/500 images (88.6\%) correctly grouped (accuracy=0.886, macro-F1=0.874)}. 
Table~\ref{tab:clustering_results} presents the confusion matrix and F1 scores.

\begin{table}[h]
\centering
\caption{Clustering results: confusion matrix and F1 scores (DINOv2 + UMAP + GMM)}
\label{tab:clustering_results}
\small
\begin{tabular}{@{}lrrrrrr@{}}
\toprule
 & \multicolumn{5}{c}{\textbf{Predicted Cluster}} & \\
\cmidrule(lr){2-6}
\textbf{Actual Species} & \textbf{Badger} & \textbf{Raccoon} & 
  \textbf{Red Fox} & \textbf{Polecat} & \textbf{Hooded} & \textbf{F1} \\
 & & \textbf{Dog} & & & \textbf{Crow} & \\
\midrule
Badger & \textbf{93} & 7 & 0 & 0 & 0 & 0.830 \\
Raccoon Dog & 31 & \textbf{61} & 4 & 0 & 4 & 0.713 \\
Red Fox & 0 & 2 & \textbf{95} & 3 & 0 & 0.914 \\
Polecat & 0 & 0 & 9 & \textbf{91} & 0 & 0.938 \\
Hooded Crow & 0 & 1 & 0 & 0 & \textbf{99} & 0.975 \\
\midrule
\textbf{Macro Average} & & & & & & \textbf{0.874} \\
\bottomrule
\end{tabular}
\end{table}

\subsection{1D Similarity Sorting}

Table~\ref{tab:sorting_results} presents coherence scores for DINOv2 + 1D 
t-SNE sorting (mean ± std over 10 runs).

\begin{table}[h]
\centering
\caption{1D t-SNE sorting result by species}
\label{tab:sorting_results}
\small
\begin{tabular}{@{}lcccc@{}}
\toprule
\textbf{Species} & \textbf{Domain} & \textbf{Coherence} & 
  \textbf{N} & \textbf{Issues} \\
\midrule
Lion (\emph{P. leo}) & Mammal & 100.0 ± 0.0\% & 100 & None \\
Giraffe (\emph{G. camelopardalis}) & Mammal & 99.2 ± 0.8\% & 100 & Oryx mix \\
Ostrich (\emph{S. camelus}) & Bird & 100.0 ± 0.0\% & 100 & None \\
Oryx (\emph{O. gazella}) & Mammal & 92.1 ± 2.3\% & 100 & Giraffe mix \\
Badger (\emph{M. meles}) & Mammal & 87.3 ± 3.1\% & 100 & Raccoon dog mix \\
Raccoon dog (\emph{N. procyonoides}) & Mammal & 85.6 ± 3.8\% & 100 & Badger mix \\
Red fox (\emph{V. vulpes}) & Mammal & 88.7 ± 2.9\% & 100 & Polecat mix \\
Polecat (\emph{M. putorius}) & Mammal & 86.2 ± 3.5\% & 100 & Fox mix \\
Pied crow (\emph{C. albus}) & Bird & 67.4 ± 4.2\% & 100 & Low-light, blur \\
Hooded crow (\emph{C. cornix}) & Bird & 69.1 ± 3.9\% & 100 & Low-light, blur \\
\midrule
\multicolumn{2}{l}{\textbf{Overall Mammals/Birds}} & \textbf{88.2 ± 1.8\%} & \textbf{1000} & \\
\midrule
Longfin batfish (\emph{A. palmaris}) & Fish & 96.3 ± 1.2\% & 100 & Minimal \\
Sixbar wrasse (\emph{C. sexfasciatus}) & Fish & 94.8 ± 1.7\% & 100 & Minimal \\
Grouper (\emph{Epinephelus} spp.) & Fish & 95.1 ± 1.4\% & 100 & Genus level \\
Barramundi (\emph{L. argentimaculatus}) & Fish & 94.2 ± 2.1\% & 100 & Minimal \\
Great barracuda (\emph{S. barracuda}) & Fish & 95.6 ± 1.3\% & 100 & Minimal \\
\midrule
\multicolumn{2}{l}{\textbf{Overall Fishes}} & \textbf{95.2 ± 0.9\%} & \textbf{500} & \\
\bottomrule
\end{tabular}
\end{table}

Visual inspection revealed that continuous 1D similarity ordering captures 
fine-grained morphological variation beyond what traditional CNNs or discrete 
clustering typically provide. 
While clustering assigns a cluster and CNNs predict fixed 
classes like "lion", the sorted sequence showed to have the capabilities of
naturally organize cropped out animal images based on more fine-grained 
biological traits, especially when they were distinct. This includes observed
patterns related to:
(1) sex, such as lion females and males appearing in distinct regions.
(2) age/maturity (antler presence/absence in deer, adult vs. juvenile lions), 
(3) individual identity (repeated sightings of the same animals clustering 
together), and (4) pose/viewpoint (giraffe legs, ostrich feet, profile vs. 
frontal views etc.).

This more nuanced structure demonstrates a critical advantage of 
continuous similarity ordering over discrete classification: rather than 
collapsing diversity into single labels, the 1D sorting preserves 
fine-grained biological variation, possibly enabling users to discover sub-species 
patterns, estimate sex ratios, track individuals, analyze population and possibly species identification. 
When combined with manual annotation workflows, this approach transforms 
zero-shot organization from mere species grouping into a tool for discovering 
ecological patterns within species.

\section{Discussion}

Analysis of misclassified images suggests clustering performance could 
substantially improve through targeted outlier removal. 
The confusion matrix reveals most errors concentrate in specific challenging 
cases: (1) extremely distant animals where morphological features are barely 
visible (e.g., black-backed jackals vs foxes at night), (2) severe motion blur 
or low-light conditions degrading image quality, (3) partial detections showing 
only body fragments (e.g., monkey tails without heads, making species 
attribution ambiguous), (4) morphologically similar species in suboptimal 
conditions. 

If systematic outlier detection methods could identify and exclude these 
problematic images, the remaining subset might achieve substantially higher 
F1 scores while still covering the majority of data. 
This suggests a promising direction: combining outlier detection with 
zero-shot clustering could make discrete clustering viable for larger species 
sets by ensuring the algorithm operates primarily on "good", 
examples rather than struggling with difficult edge cases.

Expanding beyond 5 species revealed critical scalability limitations: (1) at 10 species, 
BIC-selected component counts deviated significantly from ground truth 
(selecting 7-13 clusters instead of 10), causing F1 to drop to 0.47-0.61, 
(2) UMAP hyperparameters required tuning for optimal separation, (3) morphologically similar species 
consistently confused GMM in challenging imaging conditions.

Our GMM approach requires specifying a search range [2, 15] for the number 
of components. While BIC successfully identified the correct count (k=5) 
in our test set, this hyperparameter may need adjustment for deployments 
with substantially more or fewer species. Automated methods for determining 
appropriate search ranges remain an open question for zero-shot wildlife 
clustering.

\section{Planned Work}

We will extend evaluation to systematically address current limitations:

\textbf{Benchmark dataset creation.} create open-source evaluation datasets 
spanning multiple taxonomic categories with an increased species count. 
Document comprehensive results across different vision transformer architectures, 
clustering algorithms, and dimension reduction configurations, enabling 
reproducible comparison of zero-shot methods for wildlife applications.

\textbf{Outlier removal strategies.} Develop and evaluate systematic methods 
to detect and remove problematic images before clustering. 
Quantify performance improvements when operating on cleaned subsets versus 
full datasets, and assess whether targeted removal of challenging 
cases enables discrete clustering to scale to larger species sets.

\textbf{Multi-level clustering.} Investigate hierarchical approaches 
clustering at each taxonomic level (family → genus → species) separately, 
leveraging biological structure to reduce embedding space variance and 
improve fine-grained discrimination.

\section{Conclusion}

This work establishes baseline performance for zero-shot wildlife clustering and sorting
using vision transformers, demonstrating 88.6\% accuracy from clustering a small 
species sets and 88-95\% coherence for continuous similarity ordering in a 1D embedding space. 
Results reveal both promise and limitations: zero-shot methods excel for 
exploratory analysis and morphologically distinct species, but scalability 
challenges and morphological confusion necessitate hybrid approaches 
combining zero-shot ordering with supervised classification for production 
conservation workflows. 

The continuous 1D ordering approach deployed in Animal Detect provides 
practical value, enabling rapid dataset exploration while accelerating 
biodiversity monitoring essential for documenting climate-driven ecosystem 
changes.

\bibliographystyle{unsrtnat}
\bibliography{references}

\end{document}